# A Novel Equation based Classifier for Detecting Human in Images

Subra Mukherjee, Karen Das
Department of Electronics and Communication
Don Bosco College of Engineering and Technology
Assam Don Bosco University, Guwahati, Assam, India

## ABSTRACT

Shape based classification is one of the most challenging tasks in the field of computer vision. Shapes play a vital role in object recognition. The basic shapes in an image can occur in varying scale, position and orientation. And specially when detecting human, the task becomes more challenging owing to the largely varying size, shape, posture and clothing of human. So, in our work we detect human, based on the head-shoulder shape as it is the most unvarying part of human body. Here, firstly a new and a novel equation named as the "Omega Equation" that describes the shape of human head-shoulder is developed and based on this equation, a classifier is designed particularly for detecting human presence in a scene. The classifier detects human by analyzing some of the discriminative features of the values of the parameters obtained from the Omega equation. The proposed method has been tested on a variety of shape dataset taking into consideration the complexities of human head-shoulder shape. In all the experiments the proposed method demonstrated satisfactory results.

## General Terms
Image Processing, Pattern Classification, Machine Vision.

## Keywords
Omega Equation, Classifier, Human Detection, Omega Shape, Gaussian Mixture Model.

## 1. INTRODUCTION

Human detection in images and videos has drawn the attention of researchers worldwide in the field of machine learning. This is due to its large potential applications like visual surveillance, automated driving assistance, advanced user interface, motion based diagnosis, etc [1]. Human detection in images refers to the process of identifying the presence of human beings and distinguishing them from other non-human objects. However human detection is a very challenging and complicated task as compared to other objects due to factors such as: varying camera positions, dynamically changing background, sudden or gradual changes in illumination, varying human pose, appearance and clothing. Many of the human detection methods mostly require a high resolution direct face, or the entire body to be visible. Also some methods require an extensively huge database for matching based classification. Certain other methods detect human head by searching for circular or elliptical pattern in the image. Most of these methods seem to be insufficient if the human is very far or turning away from camera such that the face is not visible. Also sometimes the human may be partially occluded (legs) as a result of which component based detection may not be sufficient. Also matching the human head with circular or elliptical patterns leads to inaccurate detection if the images consist of some other circular shapes, as the human head shape may not be circular or elliptical all the time.

So, in this work we develop a method to overcome all these issues to a great extent. Based on observation we found that the human head-neck-shoulder region resembles very closely with the shape of the symbol **Ω** (OMEGA). In this work we have made two contributions:

- Firstly, for the first time in literature we develop a novel, new and a robust equation and we named it as *"Omega Equation"* particularly for describing the human head-neck-shoulder shape.

- Secondly, using this developed equation we design a classifier that takes into account some of the unique and discriminating features of head-neck-shoulder signature obtained from *"Omega Equation"*.

Initially in our work, we used the Euclidean distance between the unique landmark points (defined by us) in the pattern of **'S' (a parameter defined in the Omega equation)** as a measure to classify and distinguish between human and non-human objects. But after an extensive experimentation and analysis we found that Euclidean distance alone was not sufficient, so we studied a few more unique features of the pattern of **'S' for human** and finally designed a classifier to classify and detect human in images.

The main reason for detecting human using head-neck-shoulder signature is that it is the most unvarying part of human body. Our method also works well even when human is not facing the camera directly. Also the method works when the human body is partially occluded ( legs are not visible).

The rest of the paper is organized as follows: In section 2, we discuss some of the relevant work in this field; in section 3, we give a detail description of the methodology adopted for our work; in section 4 we demonstrate and discuss the results obtained experimentally followed by conclusion and future work in section5.

.

## 2. RELATED WORK

There is an extensive literature on object detection, however we discuss a few related ones in this section. A comprehensive survey of human motion analysis has been done in [1]. Although a significant amount of research has been done in this field, yet human detection is considered to be an open problem in the field of machine learning and computer vision. Human detection in images is more challenging than detecting many other objects due to several reasons like highly articulated resulting in varying shapes, pose, garments, etc [2]. Most human detection methods reported in literature can be broadly categorized into one of the three categories: motion based, shape based, component based. The non-rigid articulated human body shows a periodic property and this had been used as a strong cue for motion based classification [1] .In [3], moving human identification was done by computing some human size parameters and extracting skin segments. In shape based classification the shape information obtained from blob or silhouettes are used to classify human. Various approaches for shape based classification of objects are discussed in [4-12]. In [13] a shape based approach has been proposed where a part-template tree is matched to images hierarchically to detect human and also estimate their pose. In [14] edge-based features combined along with color and texture information was used for efficient human detection. In [15] human had been detected by detecting skin like pixels and locating each face like region. In component based human detection, various parts of the human body head, legs, right arm, left arm are detected part-wise and then the system checks to ensure that the detected components are all in proper geometric configuration and combines them using a classifier. Such kind of a work was done in [2]. Also in [16] similar kind of work was done wherein they proposed a method for human detection by modeling human as flexible assemblies of parts represented by co-occurrence of local features. In [17] part detectors were learned by boosting a number of weak classifiers based on edgelet features. Recently many shaped based techniques is seen in the literature to detect human by comparing the human head with a circular or elliptical pattern. This kind of a work was done in [18-20] where they used an SVM classifier to detect human beings based on finding people's head by searching for circular patterns through a 2D correlation using a bank of annular patterns.

## 3. SYSTEM OVERVIEW AND PROPOSED METHOD

The general flow diagram of our human detection system is as shown in figure 1:

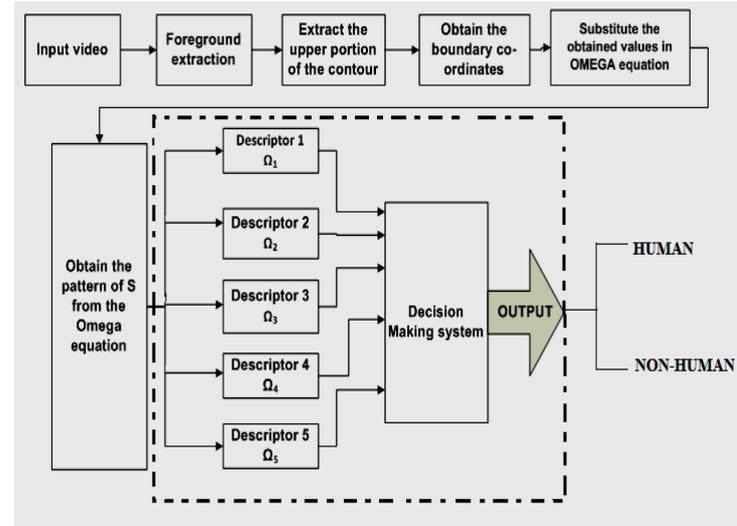

**Figure1: General flow diagram of our work**

In any human detection method the first task is to find the region of interest from the video. The region of interest refers to the moving objects in the scene. So, for this foreground extraction is done. Though a number of approaches are available, however background subtraction using Adaptive Gaussian Mixture Model is found to be the most appropriate in literature when dealing with multi-modal background.

In our prior work [21], we have performed background subtraction in real time by modeling the background using GMM (Gaussian Mixture Model). This has been used as a pre-processing step for our present work. After background subtraction the segmented foreground contours are then processed using developed Omega Equation and then finally tested for human presence with the help of the developed classifier.

### 3.1 The Omega Equation

Here work we have developed an equation named as the omega equation that describes the shape of the human head-shoulder region. The Equation is given as:

$$Y = \sqrt{(S^2 - X^2 + |X|)} - \frac{|X|}{K} \cdot \sqrt{|(S^2 - X^2)|} \quad -----(1)$$

For convenience, X is normalized to -U to +U (where U is an integer). Also, we found experimentally that 'K' generally lies between a critical range. So, keeping 'K' as constant the main idea is to find the values of 'S' for a given shape.

Let,

$$S^2 - X^2 = Q^2$$

Then eq. (1) can be written as,

$$Y + \frac{|X|}{K} Q = \sqrt{Q^2 + |X|}$$

$$Y = \sqrt{Q^2 + |X|} - \left(\frac{|X|}{K} \sqrt{Q^2}\right)$$

$$\left[\frac{X^2}{K^2} - 1\right] Q^2 + \frac{2.|X|.Y}{K} Q + [Y^2 - |X|] = 0 \quad ----- (2)$$

$$Q = \frac{-\frac{2|X|.Y}{K} \pm \sqrt{\frac{4X^2Y^2}{K^2} - 4.\left(\frac{X^2}{K^2} - 1\right)(Y^2 - |X|)}}{2\left(\frac{X^2}{K^2} - 1\right)} \quad ---- (3)$$

Let, $m^2 = X^2Y^2 - (X^2 - K^2)(Y^2 - |X|)$

$n = X^2 - K^2$

Therefore solving for S, we get,

$$S = \sqrt{\left(\frac{1}{m^2}\right)[(X^2Y^2 + m^2 \mp 2|X|.Y.m) + X^2]} \quad ------ (4)$$

Eqn. (4) gives the required solution for 'S' to implement the algorithm for human detection.

Using this equation we obtain a unique pattern of 'S' for a human contour as shown below:

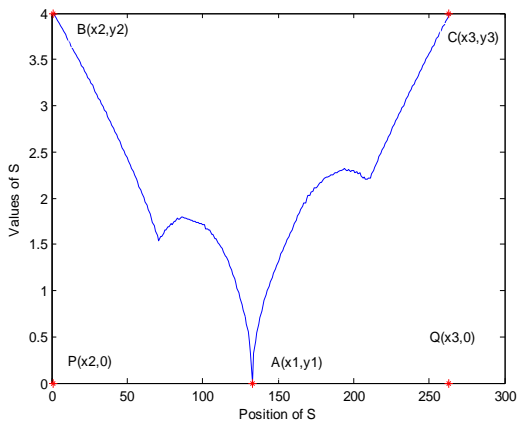

**Figure2 : Typical pattern of 'S' (obtained from omega equation) for human contour.**

We then design a classifier for human detection, by studying the unique features of this pattern.

### 3.2 Classifier for Human Detection

After background subtraction, we obtain the region of interest or the foreground contours. These are then further processed to check for the presence of human detection.

Firstly the head-neck-shoulder region of the contour is segmented and from the set of boundary points obtained, by processing the contour of the segmented objects, we then obtain the values of parameter 'S' from equation (4).

- For this, firstly we define the dimensions of the contour as shown in fig.3
  $\{Y_{max} - Y_{min}, X_{min} - X_{max}\}$

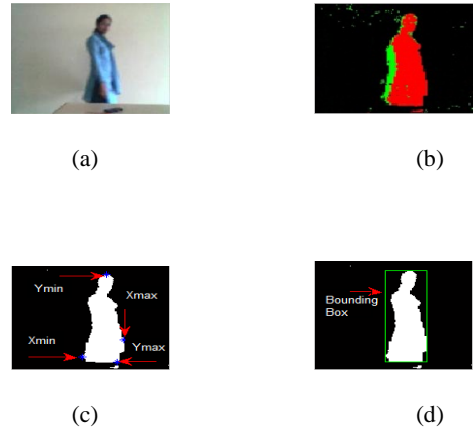

(a)       (b)

(c)       (d)

**Figure3: (a) original image, (b) image after background subtraction (c) dimensions of the image (d) Bounding box for the contour.**

- A bounding box is designed to include the object of interest and whose axes are aligned with the image axes as shown in fig.3 (d)
- Based on the set of boundary points obtained, co-ordinates ($C_x$, $C_y$) of the centroid are calculated.
- Obtain the distance d = ($C_y$ -$Y_{min}$).
- Obtain h= ½ of d, where h is the window height for extracting the head –shoulder portion of the human.
- Then we extract the set of co-ordinates from the boundary of the upper-segmented contour.
- These values are then substituted in eqn. (4) to obtain the pattern for 'S'.

After an extensive experimentation of both human and non-human contours, we observed that the values of parameter 'S' always follow a unique pattern and sequence for a human contour. Based on this fact, for each contour in a test image, we obtain the pattern for 'S' using the steps described above and then we employ an algorithm to detect the presence of human in a scene.

Firstly, the most unique feature is that the pattern of 'S' for an 'OMEGA' shape consist of only one global minima. So, we give highest priority to this, defined as descriptor 1($\Omega_1$). Then we define two more descriptors ($\Omega_2$, $\Omega_3$) based on the Euclidean distance between the landmark points A, B, C, P, Q and obtaining a threshold for human classification. Also after an extensive experimentation, we found that sometimes for certain non-human objects also the head-shoulder closely resembles an almost omega like shape contours (as shown in Table: 2 sl.no.5,7, 8). As a result of which the pattern of 'S' obtained is very closely related and symmetrical and hence the Euclidean distance between the landmark points shown in fig.3 leads to certain false detection. So, to make the system more accurate we take into consideration few more landmark points as shown in fig.4.

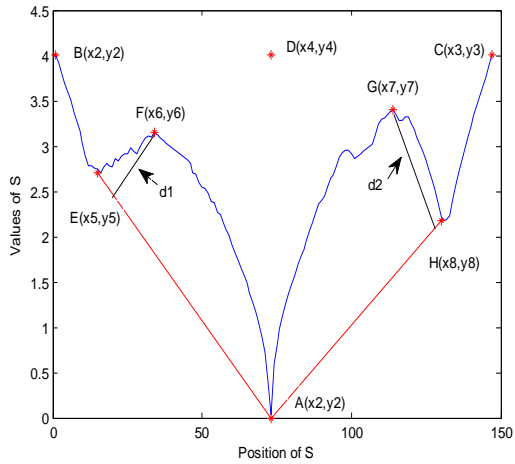

**Figure4: typical pattern of 'S' of a human, showing all the landmark points, A,B,C,E,F,G,H.**

Here we define descriptors 4 and 5 ($\Omega_4$ & $\Omega_5$) which considers the unique symmetrical pattern of the slope from points 'E' to 'F' and 'G' to 'H'. This pattern is unique and is not seen in other similar shaped objects except for human. The pattern shown in fig.4 is that of a front or back view of human. In any case, if we take the side view of human, even then we get at least one of these distances d1 or d2 lying within a threshold limit(table2:sl.no.3,4,12).

### 3.3 Classifying Algorithm

As already mentioned that the values of 'S' follows a unique pattern for a human shape, so we develop a simple but effective algorithm to classify these patterns and hence detect human based on the unanimous decision of the five descriptors. The algorithm for classifier design is as shown below:

**Table 1  Classifying algorithm**

**Algorithm:**

1. Obtain the pattern of **'S'** for each upper segmented contour.

2. Find the co-ordinates **A($x_1,y_1$)** of the global minima as shown in fig.3

3. Find co-ordinates **B($x_2, y_2$)** and **C($x_3,y_3$)** of global maxima lying to left and right of '**A**' respectively.(ref.fig3)

4. Define $\Omega_1$ (no. of global minima) =[1, for human

                            0, for non-human

5. Find the Euclidean distance between **AB** and **AC**.

6. Also find the distance **PA** and **QA**.

7. $\Omega_2$ =[ 1, if |AB-AC | > $\Omega_{2th}$

            0, otherwise

8. $\Omega_3$ =[1, if  |PA-QA|> $\Omega_{3th}$

         0,   otherwise

9. Euclidean distances **'BD'** and **'DC'** are computed,

where **'D'** is the point lying opposite to **'A'** in the line BC.

10.  $\Omega_4$= [1,  |BD-DC| < $\Omega_{4th}$

            0, otherwise]

11. Find the two local minima points '**E**' and '**H**'. Also find two local maxima points of**'** **S'** between '**x5**' and '**x2**', '**x2**' and '**x8**'.

12. Find the perpendicular distances **'d1'** and **'d2'** from '**F**' to line **'AE'** and '**G**' to line '**AH**'.

13. $\Omega_5$ = [1, if  |d1| OR |d2| <  $\Omega_{5th}$

         0, otherwise]

14. Finally take a decision:

 *If*

         {  $\Omega_1 = \Omega_2 = \Omega_3 = \Omega_4 = \Omega_5 =1$

         then  the tested contour is that of HUMAN

 *else*

            **Non- Human** }

 **Note**: $\Omega_1$=1 always for human. $\Omega_{2th}$, $\Omega_{3th}$, $\Omega_{4th}$, $\Omega_{5th}$ are the critical values of the thresholds obtained for each descriptors  experimentally.

## 4. RESULTS AND DISCUSSION

The developed human detection system has been tested on a large number of human and non-human datasets. The results obtained are satisfactory. The resolution of the camera used in the work is 120 x 160, running in a 32 bit operating system, 2.00 GHz processor, and 2 GB RAM. We have tested our developed method for a dataset containing contours of both human and non-human objects obtained after background subtraction and other pre-processing steps as mentioned above.

In the table.2, a few of our testing results are shown: In our experiment we have taken contours of human with different pose like front view, side view, back view. In front and back view as can be seen in table2, (sl. no.2, 11, 13) two local minima are obtained on either side of global minima, and as given by descriptor 5, they have a critical thresholding limit as discussed in section 3. However in case of side view like in (table2, sl.no.12) at least one local minima on either side is present. Also we have shown several other poses of human. Thus we are successfully able to detect human under varying pose at different camera angle view. Also we have tested our method for challenging animal contours like dog, bird, etc for which the pattern of 'S' that closely resemble the omega shape in certain conditions, however they do not have a local minima on either side of the global minima, and hence they were classified correctly as NON-HUMAN. Also we have tested our method for a giraffe contour (sl.no.6) which yielded two global minima in the pattern of 'S' and as such directly classified as NON-HUMAN. We have tested our results on a dataset containing 100 human contours, and 25 non-human contours. The success rate achieved is 96%. Thus it is very apparent from the results obtained that the developed method performs very well even under a challenging scenario and is very effective and robust in detecting human from images.

**TABLE2: table demonstrating the experimental results using the developed human detection system proposed in this paper.**

| NO. | Upper Segmented Contour | Plot of the Boundary | Pattern of 'S' | Decision of various descriptors ($\Omega_1, \Omega_2, \Omega_3, \Omega_4, \Omega_5$) | | | | | Output Human=1 Non-human=0 |
|---|---|---|---|---|---|---|---|---|---|
| | | | | $\Omega_1$ | $\Omega_2$ | $\Omega_3$ | $\Omega_4$ | $\Omega_5$ | |
| 1. | 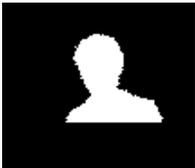 | 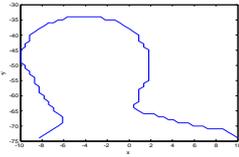 | 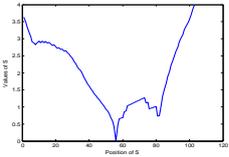 | 1 | 1 | 1 | 1 | 1 | 1 |
| 2. | 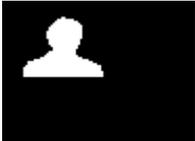 | 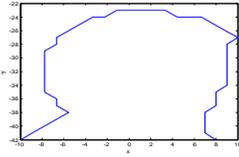 | 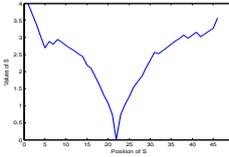 | 1 | 1 | 1 | 1 | 1 | 1 |
| 3. | 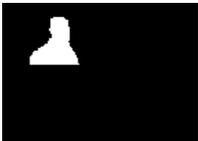 | 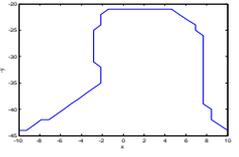 | 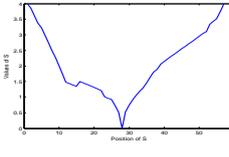 | 1 | 1 | 1 | 1 | 1 | 1 |
| 4. | 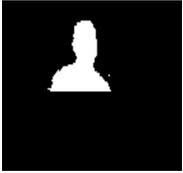 | 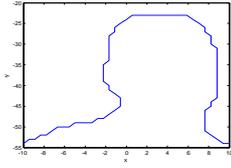 | 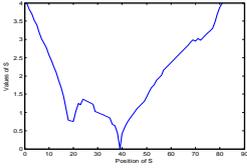 | 1 | 1 | 1 | 1 | 1 | 1 |
| 5. | 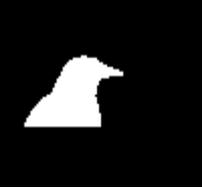 | 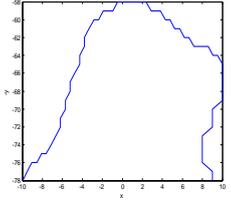 | 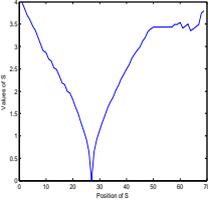 | 1 | 1 | 1 | 1 | 0 | 0 |
| 6. | 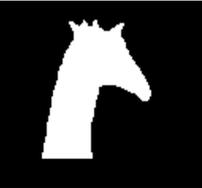 | 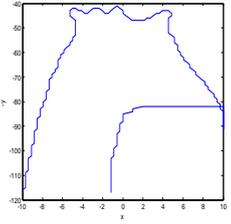 | 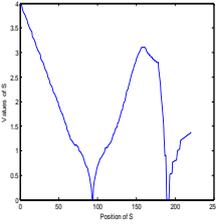 | 0 | -- | -- | -- | -- | 0 |
| 7. | 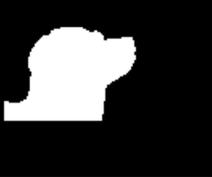 | 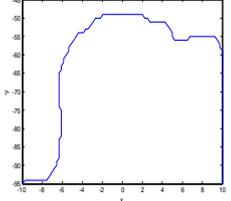 | 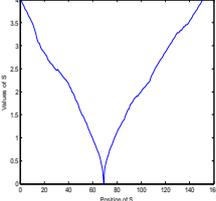 | 1 | 0 | 1 | 0 | 0 | 0 |
| 8. | 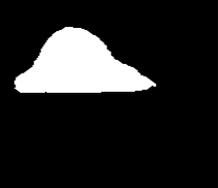 | 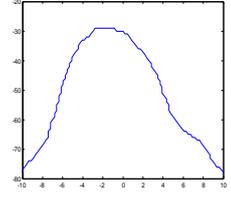 | 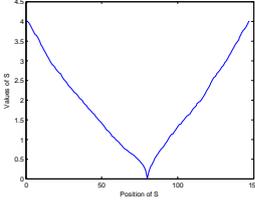 | 1 | 1 | 1 | 1 | 0 | 0 |
| 9. | 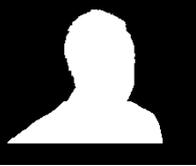 | 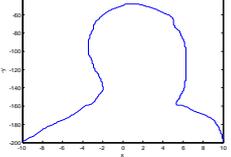 | 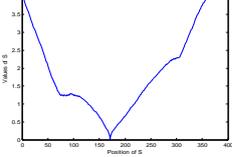 | 1 | 1 | 1 | 1 | 1 | 1 |
| 10. | 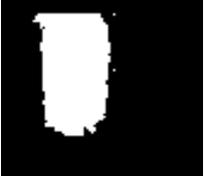 | 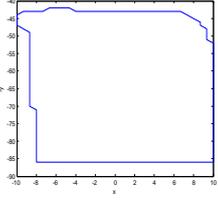 | 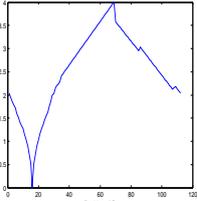 | 1 | 0 | 1 | 0 | 0 | 0 |
| 11. | 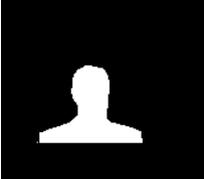 | 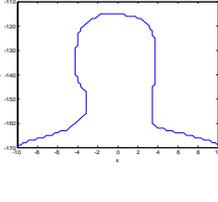 | 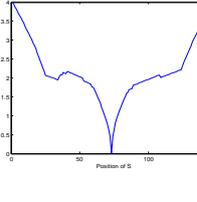 | 1 | 1 | 1 | 1 | 1 | 1 |
| 12. | 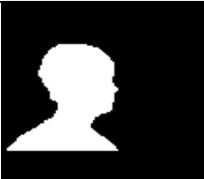 | 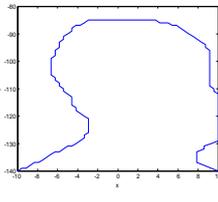 | 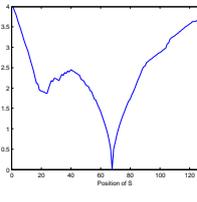 | 1 | 1 | 1 | 1 | 1 | 1 |
| 13. | 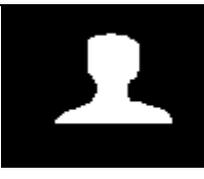 | 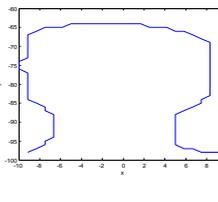 | 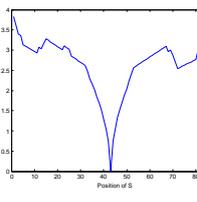 | 1 | 1 | 1 | 1 | 1 | 1 |

## 5. CONCLUSION AND FUTURE WORK

We have presented a human detection method for images based on the head- neck- shoulder signature. In this work we have made two significant contributions. Firstly we developed an equation named as "OMEGA equation" that describe the shape of human head-neck-shoulder region and then based on the pattern of the parameter 'S' obtained from equation, we have developed a classifier for human/non-human classification. In our work we have overcome the difficulties of human detection such as different pose at different camera angle views or when the human is far away or not facing the camera, partial occlusion, varying illumination condition, changing background, camouflaging clothes, etc. Detection results on several images have been presented and they validate the effectiveness of our method.

Our goal is to design a real time smart surveillance system that detects human and counts their number at every instant of time and also label some pre-defined odd situation and give threat alert if necessary. Such an autonomous system shall be of great significance to the society and shall bring an all new revolution in our present security management system specially in sensitive areas where security of human is of prime concern.

International Journal of Computer Applications (0975 – 8887)
Volume 72– No.6, May 2013